\PassOptionsToPackage{table}{xcolor}

\documentclass[manuscript]{acmart}

\setcopyright{none}
\renewcommand\footnotetextcopyrightpermission[1]{}

\usepackage{amsmath}

\usepackage{multirow}
\usepackage{makecell}
\usepackage{tabularx}
\newcolumntype{Y}{>{\centering\arraybackslash}X}

\definecolor{tableheadcolor}{RGB}{181, 82, 62}

\usepackage{pifont}

\usepackage{algorithm}
\usepackage{algorithmic}

\def\method{\texttt{RISEN}}

\title[Retrieval-guided Twin Fusion for Molecule-Text Alignment]
{Retrieval-guided Twin Fusion with Similarity-aware Contrast for Molecule-Text Alignment}

\author{Shunshun Gu}
\email{sgu82@wisc.edu}
\affiliation{
  \institution{University of Wisconsin--Madison}
  \city{Madison}
  \state{Wisconsin}
  \country{USA}
}
\author{Shengqi Qiu}
\email{sqiu53@wisc.edu}
\affiliation{
  \institution{University of Wisconsin--Madison}
  \city{Madison}
  \state{Wisconsin}
  \country{USA}
}
\author{Hang Zhou}
\email{hzh@unc.edu}
\affiliation{
  \institution{University of North Carolina at Chapel Hill}
  \city{Chapel Hill}
  \state{North Carolina}
  \country{USA}
}
\author{Xiao Luo}
\email{xiao.luo@wisc.edu}
\affiliation{
  \institution{University of Wisconsin--Madison}
  \city{Madison}
  \state{Wisconsin}
  \country{USA}
}



\begin{document}

\begin{abstract}

This paper studies the problem of molecule-text alignment, which aims to project molecules and their textual descriptions into a joint latent space for downstream tasks including molecule search and molecular property prediction. Previous approaches typically combine graph structure mining with contrastive learning to enhance joint representation learning. However, they typically neglect fine-grained semantic relationships between substructures and texts, leading to suboptimal performance on downstream tasks. Towards this end, we propose a novel approach named \underline{R}etrieval-guided Tw\underline{i}n Fu\underline{s}ion with Similarity-awar\underline{e} Co\underline{n}trast (\method{}) for molecule-text alignment. The core idea of \method{} is to construct a latent twin molecule for each substructure with cross-modal retrieval for semantic enhancement. In particular, for each substructure query, we retrieve relevant textual descriptions and sample several molecules that share similar descriptions of substructures. Then, we aggregate their representations via attention pooling for a twin latent representation, which would be further fused with the original substructure for representation enrichment. In addition, we measure the similarity across substructures and texts, which would further guide cross-modal contrastive learning with soft thresholding. Extensive experiments on benchmark datasets validate the superiority of the proposed \method{} in comparison with existing baselines. 
\end{abstract}

\maketitle

\section{Introduction}
Molecules encode rich chemical information whose systematic understanding is central to drug discovery, materials science, and molecular property prediction~\citep{wu2018moleculenet,zhang2025atomas,park2025molbridge}. Machine learning has emerged as a powerful lens for this task, with language models offering a particularly promising route, as natural-language descriptions of molecules provide functional and contextual cues that complement
structural representations~\citep{liu2023moleculestm}. Molecule--text alignment has thus become a central paradigm~\citep{edwards2021text2mol}, enabling models to jointly embed molecular structures and textual descriptions into a shared semantic space for cross-modal reasoning.

Most prior work on molecule--text alignment follows the contrastive representation learning paradigm. Early approaches such as Text2Mol \citep{edwards2021text2mol} and MoMu \citep{su2022momu} learned aligned embeddings from paired molecules and textual descriptions to support text-to-molecule retrieval. Building on this, KV-PLM \citep{zeng2022kvplm} bridged molecules and biomedical text through unified language modeling, while MoleculeSTM \citep{liu2023moleculestm} performed large-scale structure--text contrastive pre-training to obtain more general cross-modal representations that transfer to multiple downstream tasks. More recent methods pursue finer-grained correspondences. MolCA~\citep{liu2023molca} introduces a cross-modal projector to bridge graph and language spaces, while MolBridge~\citep{park2025molbridge} aligns molecular substructures with chemical phrases, further improving cross-modal learning by injecting local structural supervision.

Despite this progress, two limitations remain.
\textit{\textbf{Firstly}}, finer-grained matching can amplify noise. Real-world descriptions are often incomplete, ambiguous, and context-dependent. Expanding supervision from instance-level pairs to many substructure--phrase pairs can introduce weak or incorrect alignments, diluting useful signals and degrading generalization~\citep{huang2021learning, li2021albef}.
\textit{\textbf{Secondly}}, most methods rely on hard contrastive objectives with one-hot targets, treating all other in-batch samples as negatives~\citep{chuang2020debiased}. However, molecule--text relevance is often inherently soft, given that many molecules share scaffolds, functional groups, or pharmacophores, and treating them as negatives introduces false negatives that distort 
semantic neighborhoods. Soft contrastive targets derived from embedding similarities have shown promise in mitigating this issue in other alignment settings~\citep{park2024softcl}.

We propose \method{}, a retrieval-guided molecule--text alignment framework 
designed to address both limitations. To enrich substructure representations 
with broader chemical context that isolated fragment learning alone cannot provide, 
Retrieval-guided Twin Fusion retrieves semantically related mirror molecules for 
each substructure and aggregates them into an attention-pooled twin prototype, 
which is then fused with the original embedding to enable local motifs to inherit 
structural context from complete molecules. To mitigate false negatives arising 
from chemically similar in-batch molecules, Similarity-aware Contrast replaces 
one-hot targets with similarity-based soft targets derived from MoLFormer~\citep{ross2022molformer} 
embeddings, better preserving semantic neighborhoods in the embedding space. 
Across molecule--text retrieval and molecular property prediction benchmarks, 
\method{} achieves consistent overall gains over strong baselines. 

Our contributions are summarized as follows: \ding{182} We present a retrieval-augmented perspective for substructure-level 
molecule--text alignment, where semantically related complete molecules are 
leveraged to enrich local fragment representations with broader chemical context. \ding{183} We propose \method{}, a retrieval-guided framework that enriches 
substructure representations via twin prototype fusion and alleviates false 
negatives through similarity-aware soft targets. \ding{184} Extensive experiments on molecule--text retrieval and molecular property prediction benchmarks demonstrate consistent gains over strong baselines across both tasks.

\section{Related Work}
\label{sec:related}

\noindent\textbf{Molecule--Text Multimodal Modeling.}
Molecule--text alignment methods can be categorized by alignment granularity.
Global-level methods align complete molecular structures with full descriptions
via contrastive learning.
MoMu~\citep{su2022momu} and MoleculeSTM~\citep{liu2023moleculestm} apply
large-scale contrastive pre-training, MolCA~\citep{liu2023molca} introduces a
cross-modal projector to bridge graph and language encoders, and
instruction-tuned models such as InstructMol~\citep{cao-etal-2025-instructmol}
and BioT5+~\citep{pei-etal-2024-biot5} leverage multi-task learning for broader
generalization.
These methods operate on global molecule--text pairs and overlook fine-grained
structural correspondences.
Fine-grained methods instead capture local correspondences between molecular
fragments and chemical phrases.
\citet{yu2024multimodal} explored modality blending to encourage multi-scale
representations, \citet{min2024orma} applied optimal transport to align atom,
motif, and global levels, and Atomas~\citep{zhang2025atomas} designed a
hierarchical adaptive alignment model.
MolBridge~\citep{park2025molbridge} constructs explicit substructure--phrase
pairs via chemical fragmentation tools and applies a self-refinement mechanism
to filter noisy alignments.
Unlike these approaches, \textsc{\method{}} enriches substructure
representations with semantically related complete molecules, providing broader
chemical context without requiring explicit fragment-level annotations.

\noindent\textbf{Contrastive Learning with Soft Targets.}
Standard contrastive objectives treat all non-matching in-batch samples as
negatives, creating false negatives when semantically similar samples
co-occur~\citep{chuang2020debiased}.
\citet{li2021albef} proposed momentum-distilled soft labels to handle noisy
correspondences in vision--language pre-training, and \citet{park2024softcl}
derived soft targets from embedding similarities to preserve semantic
neighborhoods in multilingual alignment.
In the molecular domain, \citet{wang2022imolclr} demonstrated that structurally
similar molecules cause harmful false negatives and proposed fragment contrast
as mitigation.
Our Similarity-aware Contrast extends this line by constructing soft contrastive
targets from pairwise MoLFormer~\citep{ross2022molformer} embedding
similarities, directly reducing the penalty on chemically similar in-batch
molecules.

\noindent\textbf{Retrieval-Augmented Learning.}
Retrieval augmentation has been applied across NLP and vision--language modeling
to enrich representations with relevant external knowledge~\citep{lewis2020rag}.
In the molecular domain, retrieval has primarily been used at the input level
for generation tasks such as molecule
captioning~\citep{10.1109/TKDE.2024.3393356}.
\textsc{\method{}} differs by applying retrieval at the representation level
during training, aggregating retrieved mirror molecules into an attention-pooled
prototype fused with the original substructure embedding, without modifying the
inference pipeline.

\section{Methodology}
\textbf{Framework Overview.} In this paper, we introduce a retrieval-augmented alignment 
framework that extends substructure-aware contrastive learning with 
structured retrieval and soft supervision. The core idea is to enrich 
isolated substructure representations with chemical context from 
semantically related complete molecules, while replacing hard contrastive 
targets with similarity-aware soft labels to mitigate false negatives.

\subsection{Retrieval-guided Twin Fusion}
\label{sec:twin}

Substructure representations learned in isolation often lack the broader
molecular context needed to support accurate cross-modal alignment~\cite{zhao2021reducing,zhang2024semi}. To
address this, we retrieve semantically related complete molecules for each
substructure query, aggregate their embeddings into an attention-pooled
twin prototype via an attention-based mechanism, and fuse it with the
original substructure representation to provide richer chemical context.

For each substructure query, we collect the textual descriptions aligned
with it in the training set, and take all complete molecules aligned with
any of these descriptions as its mirror set. The mirror map $\mathcal{M}$ is
precomputed before training, and at each step we sample $k \leq 5$ mirror
molecules uniformly at random.

Specifically, we aggregate the embeddings of $k$ retrieved mirror
molecules $\{\mathbf{m}_1, \dots, \mathbf{m}_k\}$ into a prototype
$\mathbf{p}$ as follows:
\begin{equation}
  w_i = \frac{\exp(\mathbf{h}^{\top}\mathbf{m}_i / \sqrt{d})}
             {\sum_{j=1}^{k}\exp(\mathbf{h}^{\top}\mathbf{m}_j / \sqrt{d})},
  \quad
  \mathbf{p} = \frac{\sum_{i=1}^{k}w_i\,\mathbf{m}_i}
                    {\left\|\sum_{i=1}^{k}w_i\,\mathbf{m}_i\right\|_2}.
\end{equation}
The original representation $\mathbf{h}$ is then enriched with this 
prototype information:
\begin{equation}
  \mathbf{h}' = \mathrm{Norm}\!\left((1-\alpha)\mathbf{h} + \alpha\mathbf{p}\right),
\end{equation}
where $d$ is the embedding dimension and $\alpha$ is a fusion coefficient. This design enriches substructure representations with broader chemical context.
\subsection{Similarity-aware Contrastive Learning}
\label{sec:contrast}

Hard contrastive objectives treat all non-matching in-batch samples as 
negatives~\citep{chuang2020debiased,wang2025joint}, yet many molecules share scaffolds, functional groups, or 
pharmacophores, creating harmful false negatives. To address this, we 
leverage pairwise MoLFormer embedding similarities to construct soft 
contrastive targets that reduce the penalty on chemically similar 
in-batch molecules, better preserving semantic neighborhoods during training.

Specifically, let $S_{ij}$ denote the cosine similarity between the
normalized MoLFormer embeddings of molecules $i$ and $j$ in a mini-batch.
We apply a threshold $\tau$ to filter weak similarities and derive a soft
target distribution $\mathbf{q}^{\mathrm{soft}}$ via a temperature-scaled
softmax:
\begin{equation}
  \tilde{S}_{ij}=S_{ij}\mathbb{I}(S_{ij}\ge\tau),\;
  q_{ij}^{\mathrm{soft}}=
  \frac{\exp(\tilde{S}_{ij}/T)}
       {\sum_{l=1}^{B}\exp(\tilde{S}_{il}/T)} .
\end{equation}
For each retrieval direction, the final target distribution $\mathbf{q}$ 
is a mixture of hard labels $\mathbf{q}_{\mathrm{hard}}$ and soft labels 
$\mathbf{q}_{\mathrm{soft}}$ controlled by a mixing weight $\beta$:
\begin{equation}
  \mathbf{q} = (1-\beta)\,\mathbf{q}_{\mathrm{hard}} + \beta\,\mathbf{q}_{\mathrm{soft}}.
\end{equation}
This design yields a smoother supervision signal that better preserves semantic neighborhoods.

\subsection{Optimization Objectives}
Effective cross-modal alignment benefits from supervision that is both 
directionally balanced and robust to noise~\citep{li2021albef}. To this end, we optimize 
\method{} with two complementary objectives: a bidirectional 
contrastive loss on soft-hard mixed targets to align molecule and text 
representations, and a prototype contrastive loss to enforce alignment 
consistency between the fused representation and the twin prototype.

\noindent\textbf{Bidirectional Contrastive Loss with Soft Labels.}
Given a mini-batch of molecule-side inputs $x^{i}_{m}$ and text-side inputs
$x^{j}_{t}$, we first encode them into feature representations
$u_i = f_m(x^{i}_{m})$ and $v_j = f_t(x^{j}_{t})$.
We define the pairwise similarity $\sigma_{i,j}(u,v)$ as:
\begin{equation}
  \sigma_{i,j}(u,v)
    = \exp\!\left(\gamma\cdot\frac{u_i^{\top}v_j}{\|u_i\|_2\|v_j\|_2}\right),
\end{equation}
where $\gamma$ is a learnable logit scale.
After masking invalid pair types, and using direction-specific mixed target
distributions $q^{s\to t}_{ij}$ and $q^{t\to s}_{ij}$, the bidirectional losses
are defined as:
\begin{align}
  \mathcal{L}_{s\to t}
    &= -\frac{1}{B}\sum_{i=1}^{B}\sum_{j=1}^{B}
         q^{s\to t}_{ij}\log\frac{\sigma_{i,j}}{\sum_{l=1}^{B}\sigma_{i,l}}, \\
  \mathcal{L}_{t\to s}
    &= -\frac{1}{B}\sum_{i=1}^{B}\sum_{j=1}^{B}
         q^{t\to s}_{ij}\log\frac{\sigma_{i,j}}{\sum_{l=1}^{B}\sigma_{l,j}}.
\end{align}
The contrastive loss in this branch is the average of both directions:
$\mathcal{L}_{\mathrm{cont}} = \frac{1}{2}(\mathcal{L}_{s\to t} + \mathcal{L}_{t\to s})$.

\noindent\textbf{Total Loss.}
We optimize the contrastive loss on both the fused substructure representation
$\mathbf{h}'$ and the mirror molecule prototype $\mathbf{p}$, with an
additional classification loss $\mathcal{L}_{\mathrm{class}}$ following the
self-refinement protocol of MolBridge~\citep{park2025molbridge} to iteratively
filter unreliable alignment pairs:
\begin{equation}
  \mathcal{L}_{\mathrm{total}}
    = \underbrace{%
        \mathcal{L}_{\mathrm{cont}}(\mathbf{h}',\mathbf{t})
        + \lambda\cdot\mathcal{L}_{\mathrm{cont}}(\mathbf{p},\mathbf{t})
      }_{\mathcal{L}_{\mathrm{combined}}}
    + \mathcal{L}_{\mathrm{class}},
\end{equation}
where $\lambda$ is the prototype loss weight and $\mathbf{t}$ is the text
representation. Together, these objectives encourage \method{} to learn
representations that are semantically aligned, structurally enriched, and
robust to noisy correspondences.

\section{Experiments}

\subsection{Experimental Settings}

\noindent\textbf{Datasets.}
Following~\citet{park2025molbridge}, we train on 431,877 molecule--description pairs augmented to 2M via substructure-level alignment, and evaluate on PCDes~\citep{zeng2022kvplm} (scaffold split) and PubChem324kV2~\citep{liu2023moleculestm} for retrieval, and eight MoleculeNet~\citep{wu2018moleculenet} datasets for property prediction.

\noindent\textbf{Metrics.}
We report Recall@1/5/10/20~\citep{manning2008ir} and MRR~\citep{voorhees1999mrr} for zero-shot molecule-text retrieval. For molecular property prediction, we report ROC-AUC (\%).

\noindent\textbf{Baselines.}
We compare against representative multimodal methods, 
including MoMu~\citep{su2022momu}, MolFM~\citep{luo2023molfm}, 
MolCA~\citep{liu2023molca}, KV-PLM~\citep{zeng2022kvplm}, 
MoleculeSTM~\citep{liu2023moleculestm}, SciBERT~\citep{beltagy2019scibert}, 
Atomas~\citep{zhang2025atomas}, and MolBridge~\citep{park2025molbridge}. 
The reported baseline results are taken from \citet{park2025molbridge}.

\subsection{Empirical Results}
\label{sec:main_retrieval}

\begin{table*}[t]
\centering
  \caption{Performance comparison for zero-shot molecule-text retrieval on PCDes test set (scaffold split). Baseline results are quoted from \citet{park2025molbridge}. We \textbf{bold} the best results and \underline{underline} the second-best.}
  \label{tab:zs_retrieval}
\begin{tabularx}{\textwidth}{l c YYYY c YYYY}
\Xhline{1.2pt}
\rowcolor{tableheadcolor!20} & & \multicolumn{4}{c}{Text to Molecule} && \multicolumn{4}{c}{Molecule to Text} \\
\cline{3-6} \cline{8-11}
\rowcolor{tableheadcolor!20}\multirow{-2}{*}{Methods} & \multirow{-2}{*}{\# Params}
& R@1 & R@5 & R@10 & MRR && R@1 & R@5 & R@10 & MRR \\
\Xhline{1.2pt}
\multicolumn{11}{l}{\textbf{1D SMILES + 2D Graph}} \\
\rowcolor{gray!10}MoMu \citep{su2022momu}                       & 111M & 4.90  & 14.48 & 20.69 & 10.33 && 5.08  & 12.82 & 18.93 & 9.89  \\
MolFM \citep{luo2023molfm}                    & 138M & 16.14 & 30.67 & 39.54 & 23.63 && 13.90 & 28.69 & 36.21 & 21.42 \\
\rowcolor{gray!10}MolFM (fine-tuned) \citep{luo2023molfm}       & 138M & 29.39 & 50.26 & 58.49 & 39.34 && 29.76 & 50.53 & 58.63 & 39.56 \\
MolCA \citep{liu2023molca}                    & 111M & 35.09 & 62.14 & 69.77 & 47.33 && 37.95 & 66.81 & 74.48 & 50.80 \\
\hline
\multicolumn{11}{l}{\textbf{1D SMILES}} \\
\rowcolor{gray!10}MoleculeSTM \citep{liu2023moleculestm}        & 120M & 35.80 & --    & --    & --    && 39.50 & --    & --    & --    \\
Atomas-base \citep{zhang2025atomas}           & 271M & 39.08 & 59.72 & 66.56 & 47.33 && 37.88 & 59.22 & 65.56 & 47.81 \\
\rowcolor{gray!10}Atomas-large \citep{zhang2025atomas}          & 825M & 49.08 & 68.32 & 73.16 & 57.79 && 46.22 & 66.02 & 72.32 & 55.52 \\
MolBridge \citep{park2025molbridge}           & 155M & \underline{50.45} & \underline{70.83} & \underline{76.11} & \underline{59.63} && \underline{52.76} & \underline{73.54} & \underline{78.55} & \underline{62.25} \\
\rowcolor{gray!10}\method{} & 155M & \textbf{52.05} & \textbf{73.47} & \textbf{78.72} & \textbf{61.56} && \textbf{55.40} & \textbf{75.04} & \textbf{80.09} & \textbf{64.28} \\
\Xhline{1.2pt}
\end{tabularx}
\end{table*}

\begin{table}[t]
\centering
  \caption{Performance comparison for zero-shot molecule-text retrieval on PubChem324kV2 test set. Baseline results are quoted from \citet{park2025molbridge}. We \textbf{bold} the best results and \underline{underline} the second-best.}
  \label{tab:r1_r20}
\begin{tabularx}{\textwidth}{l YY c YY}
\Xhline{1.2pt}
\rowcolor{tableheadcolor!20} & \multicolumn{2}{c}{Text to Molecule} && \multicolumn{2}{c}{Molecule to Text} \\
\cline{2-3} \cline{5-6}
\rowcolor{tableheadcolor!20}\multirow{-2}{*}{Methods} & R@1 & R@20 && R@1 & R@20 \\
\Xhline{1.2pt}
\multicolumn{6}{l}{\textbf{1D SMILES + 2D Graph}} \\
\rowcolor{gray!10}MoMu-S \citep{su2022momu}              & 40.8 & 86.1 && 40.9 & 86.2 \\
MoMu-K \citep{su2022momu}              & 41.6 & 87.8 && 41.8 & 87.5 \\
\rowcolor{gray!10}MoleculeSTM \citep{liu2023moleculestm} & 44.3 & 90.3 && 45.8 & 88.4 \\
MolCA \citep{liu2023molca}             & 66.0 & 93.5 && 66.6 & 94.6 \\
\hline
\multicolumn{6}{l}{\textbf{1D SMILES}} \\
\rowcolor{gray!10}SciBERT \citep{beltagy2019scibert}     & 37.5 & 85.2 && 39.7 & 85.8 \\
KV-PLM \citep{zeng2022kvplm}           & 37.7 & 85.5 && 38.8 & 86.0 \\
\rowcolor{gray!10}MolBridge \citep{park2025molbridge}    & \underline{70.9} & \underline{95.6} && \underline{75.0} & \textbf{97.4} \\
\method{} & \textbf{73.7} & \textbf{96.2} && \textbf{77.2} & \underline{97.1} \\
\Xhline{1.2pt}
\end{tabularx}
\end{table}

\noindent\textbf{Zero-shot Molecule-Text Retrieval.}
Table~\ref{tab:zs_retrieval} and Table~\ref{tab:r1_r20} summarize results
on PCDes and PubChem324kV2, respectively. \method{} achieves the best performance on most metrics across both benchmarks and retrieval directions, with particularly strong gains in M2T. These results demonstrate the benefit of retrieval-enriched substructure representations over isolated fragment learning.

\begin{table*}[t]
\centering
  \caption{Results for molecular property prediction tasks (ROC-AUC\,\%) on the MoleculeNet benchmark. We \textbf{bold} the best results and \underline{underline} the second-best.}
  \label{tab:molnet}
\begin{tabularx}{\textwidth}{l YYYYYYYYY}
\Xhline{1.2pt}
\rowcolor{tableheadcolor!20}Methods & BBBP & Tox21 & ToxCast & ClinTox & MUV & HIV & BACE & SIDER & Avg. \\
\Xhline{1.2pt}
\rowcolor{gray!10}MoleculeSTM \citep{liu2023moleculestm} & 70.6 & 75.7 & 65.2 & 86.6 & 65.7 & 77.0          & 82.0          & 63.7          & 73.3 \\
MolFM \citep{luo2023molfm}             & 72.9 & 77.2 & 64.4 & 79.7 & 76.0 & 78.8          & 83.9          & 64.2          & 74.6 \\
\rowcolor{gray!10}MoMu \citep{su2022momu}                & 70.5 & 75.6 & 63.4 & 79.9 & 70.6 & 75.9          & 76.7          & 60.5          & 71.6 \\
MolCA-SMILES \citep{liu2023molca}      & 70.8 & 76.0 & 56.2 & 89.0 & --   & --            & 79.3          & 61.1          & --   \\
\rowcolor{gray!10}Atomas \citep{zhang2025atomas}          & 73.7 & 77.9 & 66.9 & 93.2 & 76.3 & \textbf{80.6} & 83.1          & 64.4          & 77.0 \\
MolBridge \citep{park2025molbridge}     & \underline{77.6} & \textbf{84.7} & \underline{70.3} & \underline{94.8} & \underline{76.8} & 77.8 & \underline{84.5} & \underline{66.9} & \underline{79.2} \\
\rowcolor{gray!10}\method{} & \textbf{78.5} & \underline{78.4} & \textbf{74.5} & \textbf{95.3} & \textbf{80.0} & \underline{80.0} & \textbf{87.8} & \textbf{67.7} & \textbf{80.3} \\
\Xhline{1.2pt}
\end{tabularx}
\end{table*}

\noindent\textbf{Molecular Property Prediction.} Following~\citet{park2025molbridge}, we evaluate on eight
MoleculeNet~\citep{wu2018moleculenet} classification datasets using the
scaffold split from DeepChem~\citep{ramsundar2019deepchem}. Table~\ref{tab:molnet} shows that \textsc{\method{}} achieves the best average ROC-AUC with gains on most tasks, confirming effective transfer to downstream property-relevant chemical semantics.

\begin{table}[t]
\centering
  \caption{Ablation study on PubChem324kV2. SL: Soft Labels, AP: Attention Pooling, ML: Mirror Loss, PF: Prototype Fusion. \textbf{Bold} = best; \underline{underline} = second-best.}
  \label{tab:ablation_pubchem}
\begin{tabularx}{\textwidth}{l YYYY c YYYY}
\Xhline{1.2pt}
\rowcolor{tableheadcolor!20} & \multicolumn{4}{c}{Text to Molecule} && \multicolumn{4}{c}{Molecule to Text} \\
\cline{2-5} \cline{7-10}
\rowcolor{tableheadcolor!20}\multirow{-2}{*}{Methods} & R@1 & R@5 & R@10 & MRR && R@1 & R@5 & R@10 & MRR \\
\Xhline{1.2pt}
\rowcolor{gray!10}\method{}        & 73.65          & \textbf{91.50} & \textbf{94.45} & \underline{81.72} && \textbf{77.15} & \textbf{93.15} & \textbf{95.75} & \textbf{84.45} \\
\method{} w/o SL & 72.85          & \textbf{91.50} & 94.15          & 80.99             && 76.65          & \underline{92.90} & \underline{95.60} & \underline{84.06} \\
\rowcolor{gray!10}\method{} w/o AP & 73.25          & 90.65          & 93.60          & 81.06             && \underline{76.90} & 92.35          & 94.85          & 83.78          \\
\method{} w/o ML & \underline{73.70} & 90.80       & 93.80          & 81.33             && 76.00          & \textbf{93.15} & 95.35          & 83.60          \\
\rowcolor{gray!10}\method{} w/o PF & \textbf{74.30} & \underline{91.15} & \underline{94.35} & \textbf{81.98} && 76.45       & 92.80          & 95.20          & 83.84          \\
MolBridge        & 70.90          & --          & --          & --             && 75.00          & --          & --          & --          \\
\Xhline{1.2pt}
\end{tabularx}
\end{table}

\noindent\textbf{Ablation Study.} To assess the contribution of each component, we conduct an ablation study on PubChem324kV2 (Table~\ref{tab:ablation_pubchem}).
Removing soft labels clearly hurts T2M R@1 and MRR, supporting their role in mitigating false negatives. Attention pooling improves both retrieval directions.
Removing mirror loss or prototype fusion yields small T2M gains but degrades M2T, suggesting that these components mainly stabilize M2T alignment.


\begin{figure}[t!]
  \centering
  \includegraphics[width=0.85\linewidth]{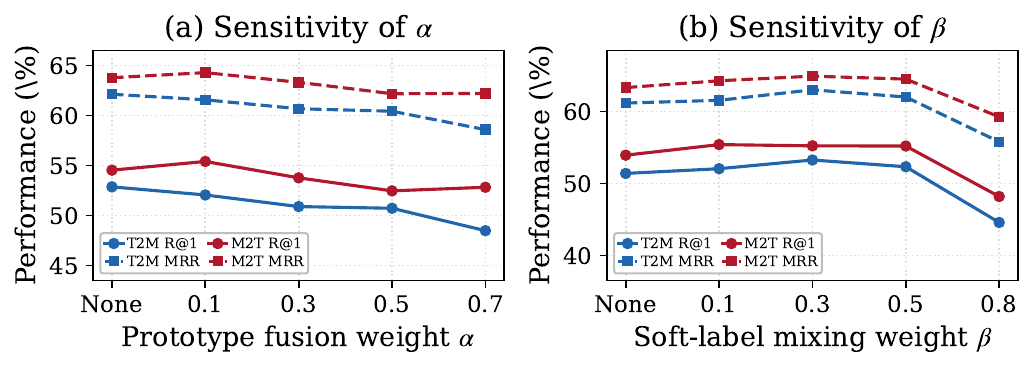}
  \caption{Sensitivity analysis of the prototype fusion weight $\alpha$
    (left) and the soft-label mixing weight $\beta$ (right) on PCDes.}
  \Description{Two line charts compare retrieval performance across prototype fusion weights and soft-label mixing weights on the PCDes benchmark.}
  \label{fig:sensitivity}
\end{figure}

\begin{figure}[t!]
  \centering
  \includegraphics[width=0.9\linewidth]{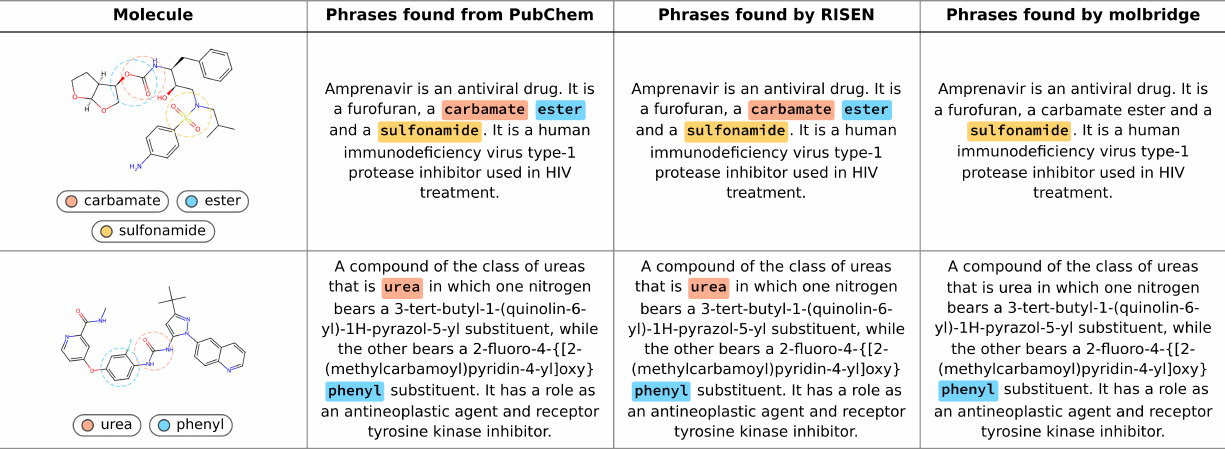}
  \caption{Qualitative comparison of functional-group phrases identified by
    \textsc{\method{}} and MolBridge~\citep{park2025molbridge} against
    ground-truth phrases from PubChem descriptions.}
  \Description{A qualitative case-study figure compares chemical functional-group phrases identified by RISEN and MolBridge against PubChem ground-truth phrases.}
  \label{fig:case_study}
\end{figure}

\noindent\textbf{Sensitivity Analysis.} We vary the prototype fusion weight $\alpha$ and the soft-label mixing
weight $\beta$ on PCDes. As shown in Figure~\ref{fig:sensitivity},
$\alpha {=}0.1$ provides the best balance across directions, indicating
that a small amount of prototype fusion suffices; $\beta{=}0.3$ yields
the strongest results, while the sharp drop at $\beta{=}0.8$ confirms
that soft labels should supplement rather than replace the original
objective.

\noindent\textbf{Case Study.}
Figure~\ref{fig:case_study} compares functional-group phrases identified by RISEN and MolBridge~\citep{park2025molbridge} against PubChem ground truth. \textsc{\method{}} recovers a more complete set, suggesting that retrieval-enriched substructure representations yield more faithful motif-level alignment.

\section{Conclusion}
We present \method{}, a retrieval-guided framework 
for molecule--text alignment that effectively enriches substructure 
representations via twin prototype fusion and 
alleviates false negatives through similarity-aware 
soft targets. Experiments on zero-shot retrieval and 
molecular property prediction benchmarks show 
consistent overall gains over strong baselines, highlighting the effectiveness of semantic retrieval and soft contrastive supervision in cross-modal molecular alignment.

\section{Limitations}
As with most prototype-based contrastive learning approaches, our method 
encodes additional mirror molecules during training, which moderately 
increases memory usage. This cost is bounded (at most k=5 per substructure 
sample) and enables structurally enriched representations that we find 
empirically beneficial, though scaling to substantially larger batch sizes 
or longer retrieval neighborhoods may pose challenges. Additionally, our
soft-label mechanism relies on similarities computed in the frozen
MoLFormer embedding space, which may not fully capture all task-relevant
chemical semantics, potentially limiting soft-label quality for
underrepresented molecular scaffolds or domain-shifted distributions.
 Finally, our evaluation focuses on molecule--text retrieval
and molecular property prediction, and extending \textsc{RISEN} to
generative tasks such as molecule captioning or text-guided molecule
generation would provide a more complete picture of its general utility.

\bibliographystyle{ACM-Reference-Format}
\bibliography{custom}

@inproceedings{park2025molbridge,
    title = "Bridging the Gap Between Molecule and Textual Descriptions via Substructure-aware Alignment",
    author = "Park, Hyuntae  and
      Kim, Yeachan  and
      Lee, SangKeun",
    editor = "Christodoulopoulos, Christos  and
      Chakraborty, Tanmoy  and
      Rose, Carolyn  and
      Peng, Violet",
    booktitle = "Proceedings of the 2025 Conference on Empirical Methods in Natural Language Processing",
    month = nov,
    year = "2025",
    address = "Suzhou, China",
    publisher = "Association for Computational Linguistics",
    url = "https://aclanthology.org/2025.emnlp-main.1197/",
    doi = "10.18653/v1/2025.emnlp-main.1197",
    pages = "23459--23479",
    ISBN = "979-8-89176-332-6"
}

@inproceedings{edwards2021text2mol,
    title = "{T}ext2{M}ol: Cross-Modal Molecule Retrieval with Natural Language Queries",
    author = "Edwards, Carl  and
      Zhai, ChengXiang  and
      Ji, Heng",
    editor = "Moens, Marie-Francine  and
      Huang, Xuanjing  and
      Specia, Lucia  and
      Yih, Scott Wen-tau",
    booktitle = "Proceedings of the 2021 Conference on Empirical Methods in Natural Language Processing",
    month = nov,
    year = "2021",
    address = "Online and Punta Cana, Dominican Republic",
    publisher = "Association for Computational Linguistics",
    url = "https://aclanthology.org/2021.emnlp-main.47/",
    doi = "10.18653/v1/2021.emnlp-main.47",
    pages = "595--607"
}

@article{liu2023moleculestm,
  title={Multi-modal molecule structure--text model for text-based retrieval and editing},
  author={Liu, Shengchao and Nie, Weili and Wang, Chengpeng and Lu, Jiarui and Qiao, Zhuoran and Liu, Ling and Tang, Jian and Xiao, Chaowei and Anandkumar, Animashree},
  journal={Nature Machine Intelligence},
  volume={5},
  number={12},
  pages={1447--1457},
  year={2023},
  publisher={Nature Publishing Group UK London}
}

@article{zeng2022kvplm,
  title={A deep-learning system bridging molecule structure and biomedical text with comprehension comparable to human professionals},
  author={Zeng, Zheni and Yao, Yuan and Liu, Zhiyuan and Sun, Maosong},
  journal={Nature communications},
  volume={13},
  number={1},
  pages={862},
  year={2022},
  publisher={Nature Publishing Group UK London}
}

@inproceedings{liu2023molca,
  title={Molca: Molecular graph-language modeling with cross-modal projector and uni-modal adapter},
  author={Liu, Zhiyuan and Li, Sihang and Luo, Yanchen and Fei, Hao and Cao, Yixin and Kawaguchi, Kenji and Wang, Xiang and Chua, Tat-Seng},
  booktitle={Proceedings of the 2023 Conference on Empirical Methods in Natural Language Processing},
  pages={15623--15638},
  year={2023}
}

@article{zhang2024semi,
  title={Semi-supervised knowledge transfer across multi-omic single-cell data},
  author={Zhang, Fan and Liu, Tianyu and Chen, Zihao and Peng, Xiaojiang and Chen, Chong and Hua, Xian-Sheng and Luo, Xiao and Zhao, Hongyu},
  journal={Advances in Neural Information Processing Systems},
  volume={37},
  pages={40861--40891},
  year={2024}
}

@article{zhao2021reducing,
  title={Reducing the covariate shift by mirror samples in cross domain alignment},
  author={Zhao, Yin and Cai, Longjun and others},
  journal={Advances in Neural Information Processing Systems},
  volume={34},
  pages={9546--9558},
  year={2021}
}

@inproceedings{wang2025joint,
  title={Joint asymmetric loss for learning with noisy labels},
  author={Wang, Jialiang and Liu, Xianming and Zhou, Xiong and Hu, Gangfeng and Zhai, Deming and Jiang, Junjun and Ji, Xiangyang},
  booktitle={2025 IEEE/CVF International Conference on Computer Vision (ICCV)},
  pages={1947--1956},
  year={2025},
  organization={IEEE}
}

@inproceedings{zhang2025atomas,
  title={Atomas: Hierarchical adaptive alignment on molecule-text for unified molecule understanding and generation},
  author={Zhang, Yikun and Ye, Geyan and Yuan, Chaohao and Han, Bo and Huang, Long-Kai and Yao, Jianhua and Liu, Wei and Rong, Yu},
  booktitle={The Thirteenth International Conference on Learning Representations},
  year={2025}
}

@misc{luo2023molfm,
      title={MolFM: A Multimodal Molecular Foundation Model}, 
      author={Yizhen Luo and Kai Yang and Massimo Hong and Xing Yi Liu and Zaiqing Nie},
      year={2023},
      eprint={2307.09484},
      archivePrefix={arXiv},
      primaryClass={q-bio.BM},
      url={https://arxiv.org/abs/2307.09484}, 
}

@article{wang2022imolclr,
  title={Improving molecular contrastive learning via faulty negative mitigation and decomposed fragment contrast},
  author={Wang, Yuyang and Magar, Rishikesh and Liang, Chen and Barati Farimani, Amir},
  journal={Journal of Chemical Information and Modeling},
  volume={62},
  number={11},
  pages={2713--2725},
  year={2022},
  publisher={ACS Publications}
}

@article{ross2022molformer,
  title={Large-scale chemical language representations capture molecular structure and properties},
  author={Ross, Jerret and Belgodere, Brian and Chenthamarakshan, Vijil and Padhi, Inkit and Mroueh, Youssef and Das, Payel},
  journal={Nature Machine Intelligence},
  volume={4},
  number={12},
  pages={1256--1264},
  year={2022},
  publisher={Nature Publishing Group UK London}
}

@article{10.1109/TKDE.2024.3393356,
  title={Empowering molecule discovery for molecule-caption translation with large language models: A chatgpt perspective},
  author={Li, Jiatong and Liu, Yunqing and Fan, Wenqi and Wei, Xiao-Yong and Liu, Hui and Tang, Jiliang and Li, Qing},
  journal={IEEE transactions on knowledge and data engineering},
  volume={36},
  number={11},
  pages={6071--6083},
  year={2024},
  publisher={IEEE}
}

@article{wu2018moleculenet,
  title={MoleculeNet: a benchmark for molecular machine learning},
  author={Wu, Zhenqin and Ramsundar, Bharath and Feinberg, Evan N and Gomes, Joseph and Geniesse, Caleb and Pappu, Aneesh S and Leswing, Karl and Pande, Vijay},
  journal={Chemical science},
  volume={9},
  number={2},
  pages={513--530},
  year={2018},
  publisher={Royal Society of Chemistry}
}

@book{manning2008ir,
  title={Introduction to information retrieval},
  author={Manning, Christopher D and Raghavan, Prabhakar and Sch{\"u}tze, Hinrich},
  year={2008},
  publisher={Cambridge university press}
}

@inproceedings{voorhees1999mrr,
  title={The TREC-8 question answering track report},
  author={Voorhees, Ellen M and others},
  booktitle={Trec},
  volume={99},
  pages={77--82},
  year={1999}
}

@book{ramsundar2019deepchem,
  title={Deep learning for the life sciences: applying deep learning to genomics, microscopy, drug discovery, and more},
  author={Ramsundar, Bharath and Eastman, Peter and Walters, Pat and Pande, Vijay},
  year={2019},
  publisher={O'Reilly Media}
}

@inproceedings{beltagy2019scibert,
  title={SciBERT: A pretrained language model for scientific text},
  author={Beltagy, Iz and Lo, Kyle and Cohan, Arman},
  booktitle={Proceedings of the 2019 conference on empirical methods in natural language processing and the 9th international joint conference on natural language processing (EMNLP-IJCNLP)},
  pages={3615--3620},
  year={2019}
}

@article{su2022momu,
  title={A molecular multimodal foundation model associating molecule graphs with natural language},
  author={Su, Bing and Du, Dazhao and Yang, Zhao and Zhou, Yujie and Li, Jiangmeng and Rao, Anyi and Sun, Hao and Lu, Zhiwu and Wen, Ji-Rong},
  journal={arXiv preprint arXiv:2209.05481},
  year={2022}
}

@article{chuang2020debiased,
  title={Debiased contrastive learning},
  author={Chuang, Ching-Yao and Robinson, Joshua and Lin, Yen-Chen and Torralba, Antonio and Jegelka, Stefanie},
  journal={Advances in neural information processing systems},
  volume={33},
  pages={8765--8775},
  year={2020}
}

@inproceedings{park2024softcl,
    title = "Improving Multi-lingual Alignment Through Soft Contrastive Learning",
    author = "Park, Minsu  and
      Choi, Seyeon  and
      Choi, Chanyeol  and
      Kim, Jun-Seong  and
      Sohn, Jy-yong",
    editor = "Cao, Yang (Trista)  and
      Papadimitriou, Isabel  and
      Ovalle, Anaelia  and
      Zampieri, Marcos  and
      Ferraro, Francis  and
      Swayamdipta, Swabha",
    booktitle = "Proceedings of the 2024 Conference of the North American Chapter of the Association for Computational Linguistics: Human Language Technologies (Volume 4: Student Research Workshop)",
    month = jun,
    year = "2024",
    address = "Mexico City, Mexico",
    publisher = "Association for Computational Linguistics",
    url = "https://aclanthology.org/2024.naacl-srw.16/",
    doi = "10.18653/v1/2024.naacl-srw.16",
    pages = "138--145"
}

@article{li2021albef,
  title={Align before fuse: Vision and language representation learning with momentum distillation},
  author={Li, Junnan and Selvaraju, Ramprasaath and Gotmare, Akhilesh and Joty, Shafiq and Xiong, Caiming and Hoi, Steven Chu Hong},
  journal={Advances in neural information processing systems},
  volume={34},
  pages={9694--9705},
  year={2021}
}

@article{huang2021learning,
  title={Learning with noisy correspondence for cross-modal matching},
  author={Huang, Zhenyu and Niu, Guocheng and Liu, Xiao and Ding, Wenbiao and Xiao, Xinyan and Wu, Hua and Peng, Xi},
  journal={Advances in Neural Information Processing Systems},
  volume={34},
  pages={29406--29419},
  year={2021}
}

@inproceedings{pei-etal-2024-biot5,
  title={Biot5+: Towards generalized biological understanding with iupac integration and multi-task tuning},
  author={Pei, Qizhi and Wu, Lijun and Gao, Kaiyuan and Liang, Xiaozhuan and Fang, Yin and Zhu, Jinhua and Xie, Shufang and Qin, Tao and Yan, Rui},
  booktitle={Findings of the Association for Computational Linguistics: ACL 2024},
  pages={1216--1240},
  year={2024}
}

@inproceedings{cao-etal-2025-instructmol,
  title={Instructmol: Multi-modal integration for building a versatile and reliable molecular assistant in drug discovery},
  author={Cao, He and Liu, Zijing and Lu, Xingyu and Yao, Yuan and Li, Yu},
  booktitle={Proceedings of the 31st International Conference on Computational Linguistics},
  pages={354--379},
  year={2025}
}

@article{lewis2020rag,
  title={Retrieval-augmented generation for knowledge-intensive nlp tasks},
  author={Lewis, Patrick and Perez, Ethan and Piktus, Aleksandra and Petroni, Fabio and Karpukhin, Vladimir and Goyal, Naman and K{\"u}ttler, Heinrich and Lewis, Mike and Yih, Wen-tau and Rockt{\"a}schel, Tim and others},
  journal={Advances in neural information processing systems},
  volume={33},
  pages={9459--9474},
  year={2020}
}

@inproceedings{yu2024multimodal,
  title={Multimodal molecular pretraining via modality blending},
  author={Yu, Qiying and Zhang, Yudi and Ni, Yuyan and Feng, Shikun and Lan, Yanyan and Zhou, Hao and Liu, Jingjing},
  booktitle={International Conference on Learning Representations},
  volume={2024},
  pages={13314--13332},
  year={2024}
}

@inproceedings{min2024orma,
  title={Exploring optimal transport-based multi-grained alignments for text-molecule retrieval},
  author={Min, Zijun and Liu, Bingshuai and Zhang, Liang and Song, Jia and Su, Jinsong and He, Song and Bo, Xiaochen},
  booktitle={2024 IEEE International Conference on Bioinformatics and Biomedicine (BIBM)},
  pages={2317--2324},
  year={2024},
  organization={IEEE}
}


\end{document}